\documentclass[journal]{IEEEtran}

\usepackage{amsmath,amssymb,amsthm}
\usepackage{graphicx}
\usepackage{booktabs}
\usepackage{array}
\usepackage{colortbl}
\usepackage{xcolor}
\usepackage{url}
\usepackage{balance}
\usepackage{multirow}
\usepackage{algorithm}
\usepackage{algorithmic}
\usepackage[colorlinks=true,
            linkcolor=blue!60!black,
            citecolor=blue!60!black,
            urlcolor=blue!60!black]{hyperref}

\definecolor{rowours}{HTML}{FEF9C3}

\newtheorem{definition}{Definition}

\title{TokenMizer: Graph-Structured Session Memory\\
for Long-Horizon LLM Context Management}

\author{Shweta~Mishra%
\thanks{Independent Researcher, India.
E-mail: \href{mailto:shweta.mishra.research@gmail.com}{shweta.mishra.research@gmail.com}.
Code and benchmark: \url{https://github.com/Shweta-Mishra-ai/tokenmizer}.
This manuscript describes TokenMizer v0.3.1.
Manuscript revised July 3, 2026.}%
}

\begin{document}
\maketitle

\begin{abstract}
Long-horizon LLM sessions outlive their context windows, and
the standard mitigations---truncation, summarization,
retrieval---share a structural flaw: they treat history as
flat text, discarding precisely the content that makes a
session resumable: decisions and their rationales, task
status, and file modification history.

We present \textbf{TokenMizer}, an open-source transparent
proxy that maintains session history as a \emph{typed
knowledge graph} and, at context boundaries, replaces the raw
transcript with a token-budgeted serialization of session
state.
The schema comprises 14 node types and 7 edge types under an
8-state lifecycle in which decisions can be superseded or
explicitly invalidated; bitemporal validity intervals support
time-travel queries; and first-class
\emph{decision-transition records} preserve why each decision
replaced its predecessor (trigger, reason, evidence).
Version 0.3.1 embeds this memory core in a production-shaped
serving layer---SSE streaming, security middleware, nine
provider adapters, a monitoring dashboard, graph exports
(D3 JSON, self-contained interactive HTML, Obsidian
Canvas)---and exposes checkpoint/resume to agents as Model
Context Protocol tools.

The evaluation is deliberately minimal and fully provenanced:
three synthetic sessions, heuristic-only extraction, one
plain-summary baseline, every value traceable to a single
versioned results file.
Graph extraction ties the baseline on task recall (75.6\%)
and exceeds it on decision recall (\textbf{85.0\%} vs.\
70.0\%) and file recall (\textbf{100\%} vs.\ 91.7\%), with
201--302-token resume blocks extracted in 8.1--529.9\,ms per
session.
At $n{=}3$ these results are directional; ceiling effects and
baseline weaknesses are analyzed explicitly.
Code, benchmark runner, and the exact results file are
released under the MIT licence.
\end{abstract}


\section{Introduction}
\label{sec:intro}

\IEEEPARstart{A}{n} iterative development session with an LLM
assistant~\cite{sweagent,copilot,housurvey} is not a sequence of
independent exchanges.
The technology chosen in turn~3 constrains the implementation
in turn~12; the error fixed in turn~7 explains the test added
in turn~15; the file created in turn~4 must stay consistent
with the schema discussed in turn~18.
Sessions accumulate \emph{structure}.

They also accumulate tokens.
Paulsen~\cite{paulsen2025} shows that models degrade on
multi-step tasks well before their advertised context limits,
and defines the Maximum Effective Context Window (MECW) as the
usable budget; for complex coding tasks it may be as low as
16{,}000 tokens against an advertised 128{,}000.
Liu et al.~\cite{lostmiddle} add that even content inside the
window is unreliable when placed mid-context.
At roughly 950~tokens per turn, a 16k MECW is exhausted in
about 16~turns---far shorter than a productive working session.

When the budget fills, something must be discarded, and the
three standard strategies all discard \emph{structure}:
truncation~\cite{brownfewshot} drops the oldest turns, which
contain precisely the goal definitions and architectural
decisions everything else depends on;
summarization~\cite{langchain,acc} produces free text that
cannot reliably distinguish ``chose Redis'' from
``considered and rejected Redis'', nor a completed task from a
pending one;
retrieval~\cite{rag,memgpt} fetches passages by embedding
similarity and can miss structurally critical but semantically
distant facts (an early runtime-version constraint, a late
deployment question).

\textbf{Position.}
TokenMizer starts from the observation that session history is
a structured knowledge artifact and should be \emph{stored} as
one.
It maintains, incrementally and transparently at a proxy
boundary, a typed knowledge graph of the session---tasks with
status, decisions with rationale and revision history, files,
errors, environment facts---and, when the context budget
fills, replaces the raw transcript with a compact,
token-budgeted serialization of that graph.
What survives a context boundary is then not ``whatever text
was recent or similar'' but the session's actual state.

\textbf{Contributions.}
All of the following ship in the released v0.3.1 codebase:
\begin{enumerate}
\item a typed session-graph schema: 14 node types, 7 edge
  types, an 8-state node lifecycle with supersession,
  invalidation, and archival, and bitemporal validity
  supporting time-travel queries
  (Section~\ref{sec:graph});
\item \emph{decision-transition records}---trigger, reason,
  evidence, confidence delta---persisted independently of
  graph pruning, turning decision history into an audit trail
  (Section~\ref{sec:transitions});
\item a hybrid extraction pipeline (deterministic heuristics
  by default, optional background LLM extraction) with a
  validation layer, and a three-tier token-budgeted checkpoint
  system (Sections~\ref{sec:extraction}--\ref{sec:checkpoint});
\item a production-shaped serving layer: transparent
  OpenAI-compatible proxy with SSE streaming, security
  middleware, nine provider adapters, compression and semantic
  caching (Section~\ref{sec:serving});
\item inspection and interoperability surfaces: monitoring
  dashboard, D3/HTML/Obsidian graph exports, and MCP
  integration exposing checkpoint/resume as agent tools
  (Section~\ref{sec:interfaces});
\item an open benchmark suite and a minimal, fully provenanced
  evaluation---3 synthetic sessions against a plain-summary
  baseline---whose every number traces to one versioned
  results file (Sections~\ref{sec:setup}--\ref{sec:results}).
\end{enumerate}

This is primarily a \emph{systems} paper.
The evaluation is intentionally small and is presented with
its weaknesses foregrounded (Section~\ref{sec:limitations});
its role is to demonstrate reproducible measurement
infrastructure and directional evidence, not to establish
production performance.

\section{Related Work}
\label{sec:related}

\textbf{Context degradation.}
The MECW concept~\cite{paulsen2025} and the
lost-in-the-middle effect~\cite{lostmiddle} jointly motivate
managing context \emph{before} overflow rather than trusting
long windows: capacity is smaller than advertised, and
mid-context content is under-attended even when it fits.

\textbf{Memory systems.}
MemGPT~\cite{memgpt} gives the LLM explicit paging control
over a hierarchical memory, targeting cross-session factual
recall; TokenMizer instead targets within-session structural
continuity, requires no model cooperation, and runs at a proxy
boundary with no application changes.
LangChain's ConversationKGMemory~\cite{langchain} is the
closest architectural relative, but its graph is untyped and
schema-free, with lifecycle management left to the developer;
TokenMizer enforces a typed schema, monotonic status
transitions, and validation.
Active Context Compression~\cite{acc} spends LLM calls to
compress in-session context; TokenMizer's default path is
deterministic and inference-free.

\textbf{Prompt compression.}
LLMLingua and LongLLMLingua~\cite{llmlingua,longllmlingua}
drop low-importance tokens using a small proxy model;
RECOMP~\cite{recomp} selectively summarizes retrieved
passages.
These operate on \emph{text}; TokenMizer's resume block is
generated from \emph{graph state}, which is what enables typed
queries (``which decisions concern authentication?'') and
status-aware resumes.
The compression engine can optionally invoke LLMLingua-family
stages for large payloads (Section~\ref{sec:serving}).

\textbf{Graph-based representation.}
GraphRAG~\cite{graphrag} builds document-corpus graphs for
retrieval; its schema has no notion of task lifecycle, error
resolution, or decision revision, which are the core of a
\emph{session} graph.

\textbf{Model Context Protocol.}
MCP~\cite{mcp} standardizes tool discovery and invocation for
LLM clients.
TokenMizer uses it in the opposite of the usual direction:
rather than giving the model access to external data, it gives
agents access to \emph{their own session memory}
(Section~\ref{sec:mcp}).
Table~\ref{tab:related} summarizes the comparison.

\begin{table}[t]
  \centering
  \caption{Qualitative comparison. Latency entries for other
  systems are order-of-magnitude characterizations from their
  descriptions; the TokenMizer entry is the measured
  per-session heuristic extraction range on the 3-session
  benchmark (Table~\ref{tab:v3results}).}
  \label{tab:related}
  \renewcommand{\arraystretch}{1.25}
  \begin{tabular}{@{}lp{1.2cm}p{1.3cm}p{1.1cm}r@{}}
    \toprule
    \textbf{System} & \textbf{Struct. graph} &
    \textbf{Mgmt. latency} &
    \textbf{Transparent proxy} & \textbf{API cost} \\
    \midrule
    Sliding window      & None      & $<$1\,ms   & Yes & \$0  \\
    LangChain KG Mem.   & Untyped   & $>$200\,ms & No  & Yes  \\
    MemGPT              & Vector DB & 50--200\,ms& No  & Yes  \\
    ACC                 & None      & $>$100\,ms & No  & Yes  \\
    RECOMP              & None      & 200\,ms    & No  & Yes  \\
    \rowcolor{rowours}
    \textbf{TokenMizer v0.3.1}
      & \textbf{Typed, 14-class}
      & \textbf{8--530\,ms /session ($n{=}3$)}
      & \textbf{Yes}   & \textbf{\$0 (heuristic)} \\
    \bottomrule
  \end{tabular}
\end{table}

\section{Problem Statement}
\label{sec:problem}

\begin{definition}[Session and overflow]
A session $S = \langle m_1, \ldots, m_n \rangle$ is an ordered
message sequence with token counts $\tau(m_i)$ and cumulative
count $T(t) = \sum_{i \le t} \tau(m_i)$.
The session overflows at the first turn $n^*$ with
$T(n^*) > T_{\mathrm{MECW}}$.
For $\bar{\tau} = 950$ and $T_{\mathrm{MECW}} = 16{,}000$:
$n^* = 16$.
\end{definition}

\begin{definition}[Resume block]
A resume block $R$ is a text serialization of the session
graph $G$ within a budget $T_{\mathrm{budget}}$.
After a checkpoint at turn $n^*{-}2$, the context becomes
$C' = R \cup \{m_{n^*-1}, m_{n^*}\}$ with
$|C'| \le T_{\mathrm{MECW}}$ by construction.
\end{definition}

\begin{definition}[Recall and information loss]
For entity category $c \in \{\mathrm{task}, \mathrm{dec},
\mathrm{file}\}$, recall $\mathrm{Rec}_c$ is the fraction of
ground-truth entities fuzzy-matched
(Definition~\ref{def:fuzzy}) by at least one extracted label.
Information loss is
\begin{equation}
  \mathcal{L} = 1 - \tfrac{1}{3}\bigl(\mathrm{Rec}_\mathrm{task}
   + \mathrm{Rec}_\mathrm{dec} + \mathrm{Rec}_\mathrm{file}\bigr).
  \label{eq:infoloss}
\end{equation}
Equal weighting is a simplifying assumption
(Section~\ref{sec:limitations}).
\end{definition}

\begin{definition}[Token efficiency]
$\eta = \frac{1 - \mathcal{L}}{|R| / 100}$: mean recall per
100 resume tokens.
\label{def:eff}
\end{definition}

The design goal is a representation that maximizes
recall---especially of decisions, whose loss is the most
damaging---at a small, budget-controlled $|R|$.

\section{Design: The Session Graph}
\label{sec:design}

\begin{figure}[t]
  \centering
  \includegraphics[width=\columnwidth]{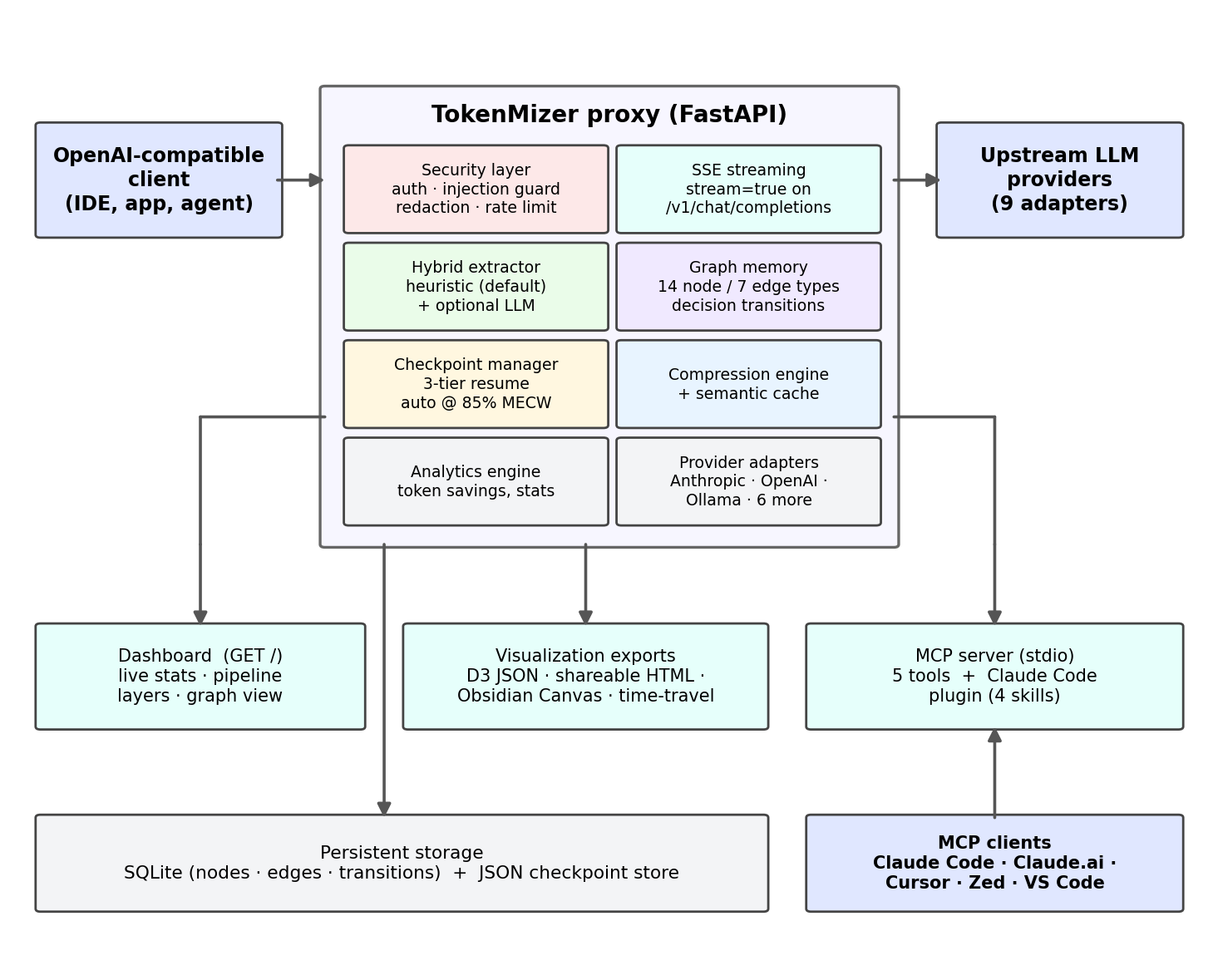}
  \caption{TokenMizer v0.3.1. The FastAPI proxy interposes
    between OpenAI-compatible clients and nine provider
    adapters. In-proxy components: security layer, SSE
    streaming, hybrid extractor, graph memory, checkpoint
    manager, compression + semantic cache, analytics.
    Read-side surfaces: dashboard, visualization exports, and
    an MCP server (with Claude Code plugin) used directly by
    MCP clients. State persists in SQLite plus a JSON
    checkpoint store.}
  \label{fig:arch}
\end{figure}

Figure~\ref{fig:arch} shows the v0.3.1 architecture.
This section describes the memory core; Sections~\ref{sec:serving}
and~\ref{sec:interfaces} describe the serving layer and the
inspection surfaces.
Module and endpoint names are given throughout so that every
claim can be checked against the repository.

\subsection{Typed Nodes and Edges}
\label{sec:graph}

\begin{figure}[t]
  \centering
  \includegraphics[width=\columnwidth]{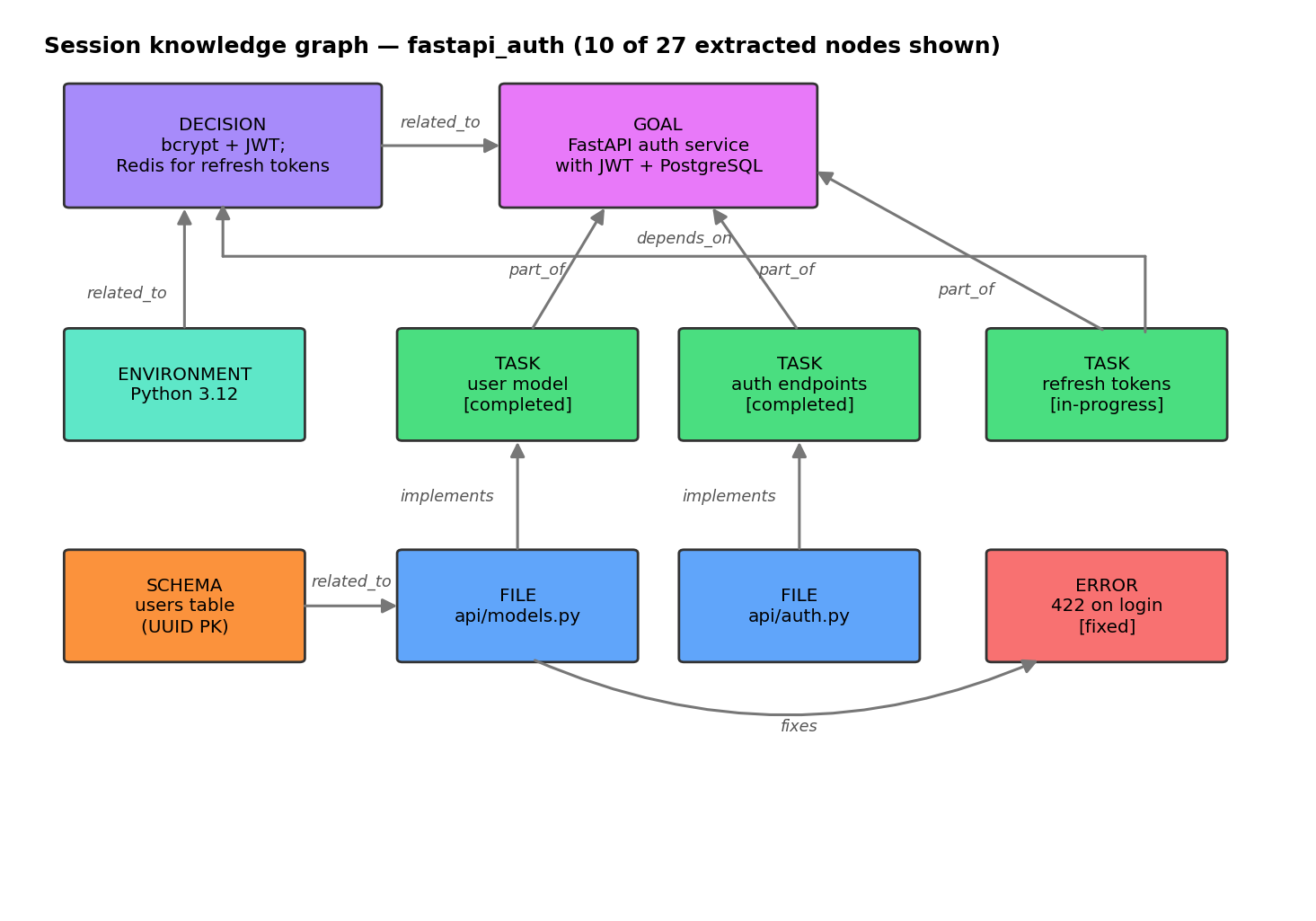}
  \caption{Fragment of the extracted session graph for the
    \texttt{fastapi\_auth} benchmark session (10 of 27 nodes).
    The GOAL anchors the hierarchy; TASK nodes carry status;
    the DECISION node records the bcrypt+JWT/Redis choice that
    the in-progress refresh-token task depends on; the FILE
    node that fixed the 422 error is linked to it by a typed
    \textsc{fixes} edge.}
  \label{fig:kgraph}
\end{figure}

The schema (\texttt{graph\_memory/types.py}) defines 14 node
types in three functional groups---%
\emph{action} nodes carrying status
(\texttt{TASK}, \texttt{FILE}, \texttt{ERROR}, \texttt{TEST},
\texttt{SCHEMA}),
\emph{decision} nodes
(\texttt{DECISION}, \texttt{DEPENDENCY}, \texttt{API},
\texttt{ENDPOINT}),
and \emph{context} nodes
(\texttt{GOAL}, \texttt{ENVIRONMENT}, \texttt{PROJECT},
\texttt{CONCEPT}, \texttt{AGENT})---%
connected by 7 directed edge types
(\texttt{DEPENDS\_ON}, \texttt{RELATED\_TO},
\texttt{IMPLEMENTS}, \texttt{FIXES}, \texttt{BLOCKS},
\texttt{PART\_OF}, \texttt{SUPERSEDES}).
Figure~\ref{fig:kgraph} shows a fragment of the graph
extracted from one benchmark session.

Every node carries a normalized label, an importance score
(reinforced on re-reference, decayed with inactivity; governs
serialization priority and pruning), a validator-assigned
confidence, timestamps, and a validity interval
(Section~\ref{sec:bitemporal}).
Deduplication is identity-based over normalized label and
type, so re-extraction updates rather than duplicates.
Pruning evicts low-importance nodes above the configured
graph-size cap, but \texttt{DECISION}, \texttt{GOAL},
\texttt{ENVIRONMENT}, and \texttt{SCHEMA} nodes are
unconditionally retained: they encode session-defining
context.

\subsection{The 8-State Lifecycle}
\label{sec:lifecycle}

\begin{figure}[t]
  \centering
  \includegraphics[width=\columnwidth]{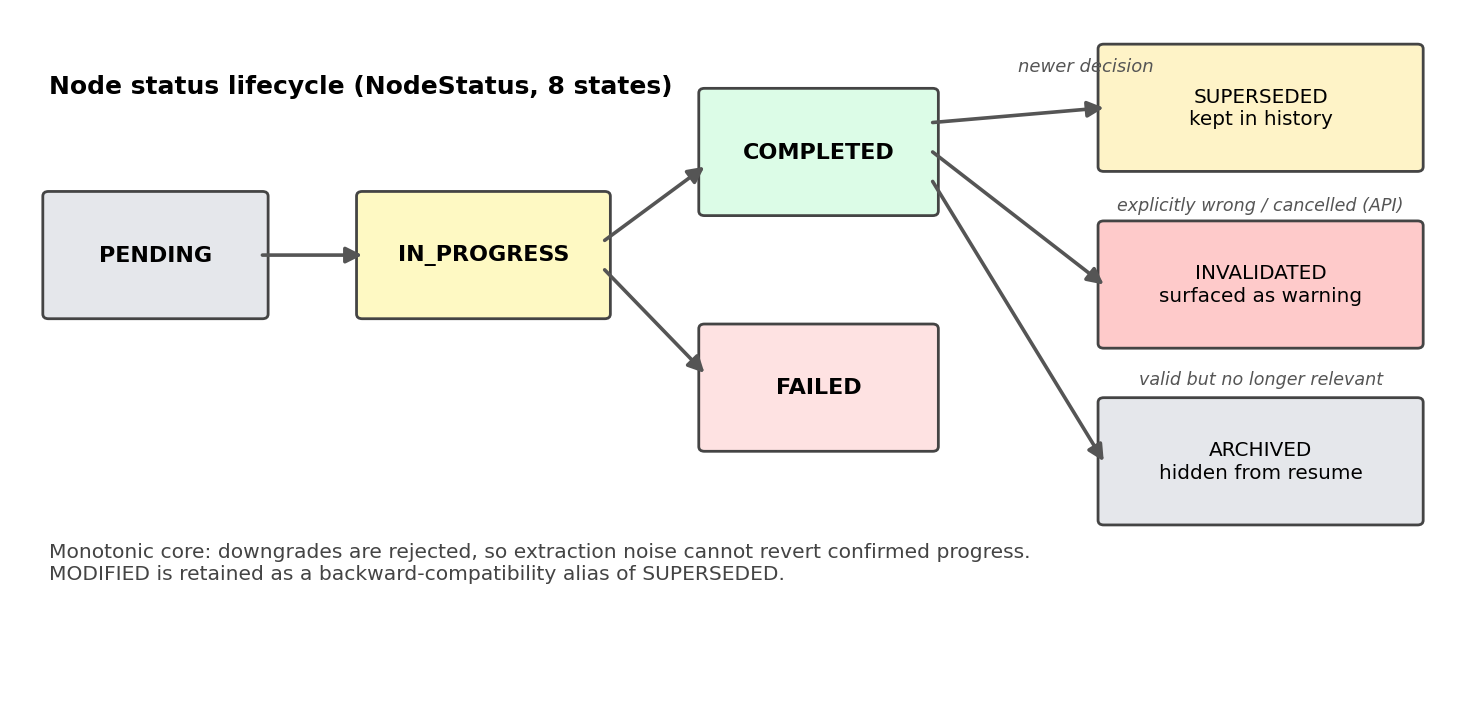}
  \caption{Node status lifecycle. The monotonic core
    (left) prevents status downgrades; the three revision
    states (right) let decision history survive without
    polluting the active resume: superseded decisions stay in
    history, invalidated ones become warnings, archived ones
    disappear from resumes but not from the graph.}
  \label{fig:lifecycle}
\end{figure}

Long sessions do not merely complete work; they \emph{revise}
it.
v0.3.1 therefore extends the classic four-state task lifecycle
with three revision states (Figure~\ref{fig:lifecycle}):
\texttt{SUPERSEDED} (replaced by a newer decision, kept in
history), \texttt{INVALIDATED} (explicitly wrong or cancelled,
surfaced as a \emph{warning} in future resumes so the model
does not re-adopt a rejected choice), and \texttt{ARCHIVED}
(valid but no longer relevant, hidden from resumes).
\texttt{MODIFIED} persists as a backward-compatibility alias.
The core transitions are monotonic---downgrades are
rejected---so extraction noise cannot revert confirmed
progress.
Invalidation is user-triggerable via
\texttt{POST /api/decision/invalidate}.

\subsection{Bitemporal Validity and Time Travel}
\label{sec:bitemporal}

Each node records \texttt{valid\_from} and
\texttt{valid\_until} (open interval = currently valid), so
the graph stores not only what is known but \emph{when it was
true}.
\texttt{GET /api/graph/\{id\}/history?at\_time=$t$}
reconstructs the graph as of time $t$: ``what did the session
believe before the database decision was reversed?'' is a
supported query, not a log-archaeology exercise.

\subsection{Decision-Transition Records}
\label{sec:transitions}

When one decision supersedes another, v0.3.1 stores a
\texttt{DecisionTransition} record: the \emph{trigger} (which
message caused the change), the \emph{reason} (why the old
choice became wrong), the \emph{evidence} (a direct quote),
and a confidence delta.
These records live in their own SQLite table---outside the
node/edge store---for two deliberate reasons: they must
survive graph pruning, and they are independently queryable
(\texttt{GET /api/graph/\{id\}/transitions}).
The effect is that a resume can say not merely ``using
SQLite'' but ``switched from PostgreSQL to SQLite because
hosting cost was raised in message~12''---the piece of
context whose absence most reliably causes a resumed session
to re-litigate settled questions.

\subsection{Hybrid Extraction and Validation}
\label{sec:extraction}

The default extractor
(\texttt{graph\_memory/hybrid\_extractor.py}) is a compiled
regular-expression pipeline over task, decision, file, error,
environment, and endpoint trigger patterns, with
compound-sentence and CSV-decision splitting.
It makes no inference calls; its measured cost is
8.1--529.9\,ms per full session on the benchmark corpus
(Section~\ref{sec:results}).
An optional LLM extraction path sends recent messages to a
small model with a structured JSON prompt; on the proxy path
it runs as a \emph{background} task after the response
returns, adding no user-visible latency.
It targets the implicit phrasing (``let's go with Redis'')
that regexes miss, and is \emph{not} evaluated in this paper.

Candidate nodes pass a validator that scores confidence,
rejects candidates below a configurable floor (default 0.50),
and applies type correction (extension-matched labels
$\to$ \texttt{FILE}; URL patterns $\to$ \texttt{ENDPOINT}).
Scoring weights are heuristic design choices, not learned
parameters.

\subsection{Checkpoints and Resume Tiers}
\label{sec:checkpoint}

Checkpoints trigger automatically at 85\% of the configured
MECW, or manually via REST
(\texttt{POST /api/checkpoint}), CLI, or the
\texttt{checkpoint\_session} MCP tool.
Serialization walks nodes in importance order into three
budgeted tiers: \emph{critical} ($\le$100 tokens: goal,
in-progress tasks, top decision), \emph{standard}
($\le$300: all tasks, decisions, files, environment), and
\emph{full} ($\le$600: complete graph).
Resumes are retrievable per tier via
\texttt{GET /api/resume/\{id\}?level=\ldots}.
Each checkpoint also stores a graph \emph{diff} against its
predecessor, so repeated checkpointing costs $O(\Delta)$, not
$O(|G|)$.

\section{Serving Layer}
\label{sec:serving}

\subsection{Transparent Proxy and Streaming}
TokenMizer implements the OpenAI Chat Completions interface
(\texttt{POST /v1/chat/completions}); clients change only
\texttt{base\_url}.
A \texttt{session\_id} field in the request body activates the
pipeline; without it, requests pass through untouched, making
adoption per-session and reversible.
Nine provider adapters are included (Anthropic, OpenAI,
DeepSeek, Mistral, OpenRouter, Grok, Cohere, Gemini, Ollama).

v0.3.1 streams: \texttt{stream=true} yields a
\texttt{text/event-stream} response relaying the provider's
native streaming interface (implemented natively for the
Anthropic, OpenAI-compatible, and Ollama adapters; providers
without native support return an explicit error rather than
degrading silently).
Three design points matter.
Semantic-cache hits stream as a single chunk, preserving a
uniform client interface.
Graph extraction and analytics run \emph{after} the stream
closes, so memory management never delays time-to-first-token.
Output trimming is disabled mid-stream---emitted tokens cannot
be retracted---so compression guarantees differ slightly
between the streaming and non-streaming paths.

\subsection{Security Middleware}
State-mutating endpoints require API-key authentication; the
chat path adds a prompt-injection guard and rate limiting.
Credential redaction (API keys, private keys, password
patterns $\to$ \texttt{[REDACTED]}) runs \emph{once at
ingress}, before any downstream consumer---extractor, cache,
provider call, analytics---sees the content; downstream paths
are therefore safe by construction rather than by repeated
defensive filtering.

\subsection{Compression and Semantic Cache}
The compression engine applies staged heuristics to large
inputs (AI-filler removal, order-preserving deduplication,
whitespace normalization, language-aware comment stripping,
history pruning with a protected recent window, file-type
aware truncation), with optional
LLMLingua-family~\cite{llmlingua,longllmlingua} neural stages
behind a semantic-similarity quality gate; an output trimmer
reduces verbose responses on the non-streaming path.
The semantic cache stores responses keyed by
sentence-transformer embeddings~\cite{sentbert} with cosine
threshold (default 0.92), LRU eviction, and TTL.

\textbf{Scope note}: the v0.3.1 results file contains no
compression or cache measurements, so this paper makes no
quantitative claims about either subsystem; benchmarking them
is planned work (Section~\ref{sec:limitations}).

\section{Inspection and Interoperability}
\label{sec:interfaces}

A memory system that silently accumulates beliefs about a
session invites justified distrust.
v0.3.1 therefore makes the graph an \emph{inspectable
artifact} through three surfaces.

\subsection{Dashboard and Analytics}
\texttt{GET /} serves a self-contained monitoring dashboard:
live session statistics, token-savings figures from the
analytics engine (\texttt{analytics/engine.py}, also exposed
at \texttt{GET /api/stats}), per-layer pipeline status,
node-type distribution, and an interactive graph view.
\texttt{GET /health} supports container orchestration.

\subsection{Visualization Exports}
\label{sec:viz}
Three exports serve different audiences
(\texttt{graph\_memory/visualization.py}):
\textbf{D3 JSON} (\texttt{/api/graph/\{id\}/viz})---nodes and
edges with per-type colors and sizes and per-\emph{status}
opacity, so superseded decisions visibly fade and invalidated
ones are nearly transparent;
\textbf{shareable HTML} (\texttt{/api/graph/\{id\}/html})---a
fully self-contained interactive force-layout page requiring
no server once saved, suitable for design reviews;
\textbf{Obsidian Canvas}
(\texttt{/api/graph/\{id\}/obsidian})---a \texttt{.canvas}
file that opens directly in an Obsidian vault, bridging
session memory into personal knowledge management.
Combined with time-travel (Section~\ref{sec:bitemporal}), a
user can audit what the system believes and when it started
believing it.

\subsection{MCP Integration}
\label{sec:mcp}
A stdio MCP server (\texttt{tokenmizer/mcp/server.py}) exposes
five tools---\texttt{checkpoint\_session},
\texttt{resume\_session}, \texttt{get\_graph\_stats},
\texttt{analyze\_file}, \texttt{get\_savings\_stats}---to any
MCP client (Claude Code, Claude.ai, Cursor, Zed, VS Code).
The server is a thin client of the HTTP API, so MCP and REST
observe identical state.
A Claude Code plugin (\texttt{.claude-plugin/}, with a
marketplace manifest) layers four skills
(\texttt{checkpoint}, \texttt{resume}, \texttt{analyze},
\texttt{stats}) over these tools.
The rationale: long-horizon agents are exactly the clients
that overflow context, and MCP lets the agent itself decide
when to checkpoint (before a task switch) and when to resume
(on re-entering a project), complementing the proxy's
automatic threshold.

\section{Implementation}
\label{sec:impl}

TokenMizer v0.3.1 is $\sim$9{,}500 lines of Python 3.10+
across \texttt{api}, \texttt{graph\_memory},
\texttt{checkpoints}, \texttt{compression},
\texttt{semantic\_cache}, \texttt{providers},
\texttt{security}, \texttt{mcp}, \texttt{dashboard},
\texttt{analytics}, \texttt{filters}, \texttt{state},
\texttt{config}, and \texttt{core}.
It uses FastAPI~\cite{fastapi} with Uvicorn,
\texttt{tiktoken}~\cite{tiktoken} (\texttt{cl100k\_base}) for
token accounting, and SQLite~\cite{sqlite} for nodes, edges,
transitions, and checkpoints.
The test suite contains 220 test functions across unit,
integration, chaos-recovery, and memory-accuracy categories.
All operational thresholds (MECW percentage, confidence floor,
cache similarity, compression minimum) are YAML-configurable.

\section{Evaluation Setup}
\label{sec:setup}

\subsection{Provenance and Scope}
Every quantitative value below comes from one file,
\texttt{benchmarks/results/results\_v3\_code-0.3.1.json},
generated on 2026-07-03 by the released runner
(\texttt{benchmarks/checkpoint\_accuracy/runner\_v2.py})
against v0.3.1 on Windows~11 under Python~3.14, and
regenerable with:
\begin{small}
\begin{verbatim}
python -m benchmarks.checkpoint_accuracy.runner_v2
\end{verbatim}
\end{small}
The scope is deliberately minimal---\textbf{3 synthetic
sessions, heuristic-only extraction, one baseline}---a
smoke-level, fully reproducible measurement of the released
code, not an empirical study.
Results from earlier code versions (including a prior
21-session corpus) are not cited: they do not reflect the
v0.3.1 pipeline.
The repository also ships \texttt{graph\_retrieval} and
\texttt{latency} benchmark runners whose results are not yet
persisted to versioned files.

\subsection{Corpus and Ground Truth}
Table~\ref{tab:corpus} lists the three scripted developer
conversations.
All use explicit imperative phrasing (``Decided:~\ldots'',
``Completed:~\ldots'', ``Fixed:~\ldots'')---a style favorable
to heuristic extraction; the resulting ceiling effects are
discussed in Section~\ref{sec:limitations}.
Each session is annotated by the author with completed tasks,
pending tasks, decisions, and files.

\begin{table}[t]
  \centering
  \caption{Benchmark corpus (3 synthetic sessions). GT:
  ground-truth counts (completed tasks / decisions / files).}
  \label{tab:corpus}
  \renewcommand{\arraystretch}{1.2}
  \begin{tabular}{@{}lp{2.6cm}rr@{}}
    \toprule
    \textbf{Session} & \textbf{Domain} & \textbf{Turns} &
    \textbf{GT (T/D/F)} \\
    \midrule
    \texttt{fastapi\_auth}    & Backend auth service & 18 & 5/5/5 \\
    \texttt{react\_dashboard} & Frontend dashboard   & 10 & 3/5/4 \\
    \texttt{ml\_pipeline}     & ML training pipeline & 10 & 3/4/4 \\
    \bottomrule
  \end{tabular}
\end{table}

\subsection{Metrics and Matching}
Recall per category and information loss follow
Eq.~\eqref{eq:infoloss}; resume size is the token count of the
standard-tier block; latency is wall-clock full-session
heuristic extraction.
Matching uses the runner's fuzzy protocol:

\begin{definition}[Fuzzy match]
\label{def:fuzzy}
Labels $a, b$ match iff $a \subseteq b$, or $b \subseteq a$,
or $\frac{|W_a \cap W_b|}{\min(|W_a|,|W_b|)} \ge 0.50$ where
$W_x$ is the set of $\ge$3-character tokens of $x$.
\end{definition}

\noindent
Recall denominates over ground truth (Appendix~A gives the
exact code).
The identical protocol scores both methods, so matching
leniency favors neither.

\subsection{Baseline}
\label{sec:baseline}
The comparison is a \textbf{plain-summary baseline}: a keyword
scan over the concatenated transcript---technology-name
matching for decisions, path patterns for files,
completion-marker snippets for tasks---approximating what a
naive ``summarize this conversation'' surfaces.
Two caveats are stated up front because both \emph{favor the
baseline}: it reads the full transcript with no token budget,
and its technology-keyword list is hand-specified with
substantial overlap with the ground-truth decision vocabulary.
Stronger baselines (LLM summaries, MemGPT-style memory,
retrieval) are future work.

\section{Results}
\label{sec:results}

\textbf{Reading guidance.}
With $n{=}3$ synthetic sessions, every number below is
directional evidence about the released pipeline, not an
estimate of production performance; no statistical testing is
meaningful at this sample size and none is claimed.

\subsection{Recall}
\begin{figure}[t]
  \centering
  \includegraphics[width=\columnwidth]{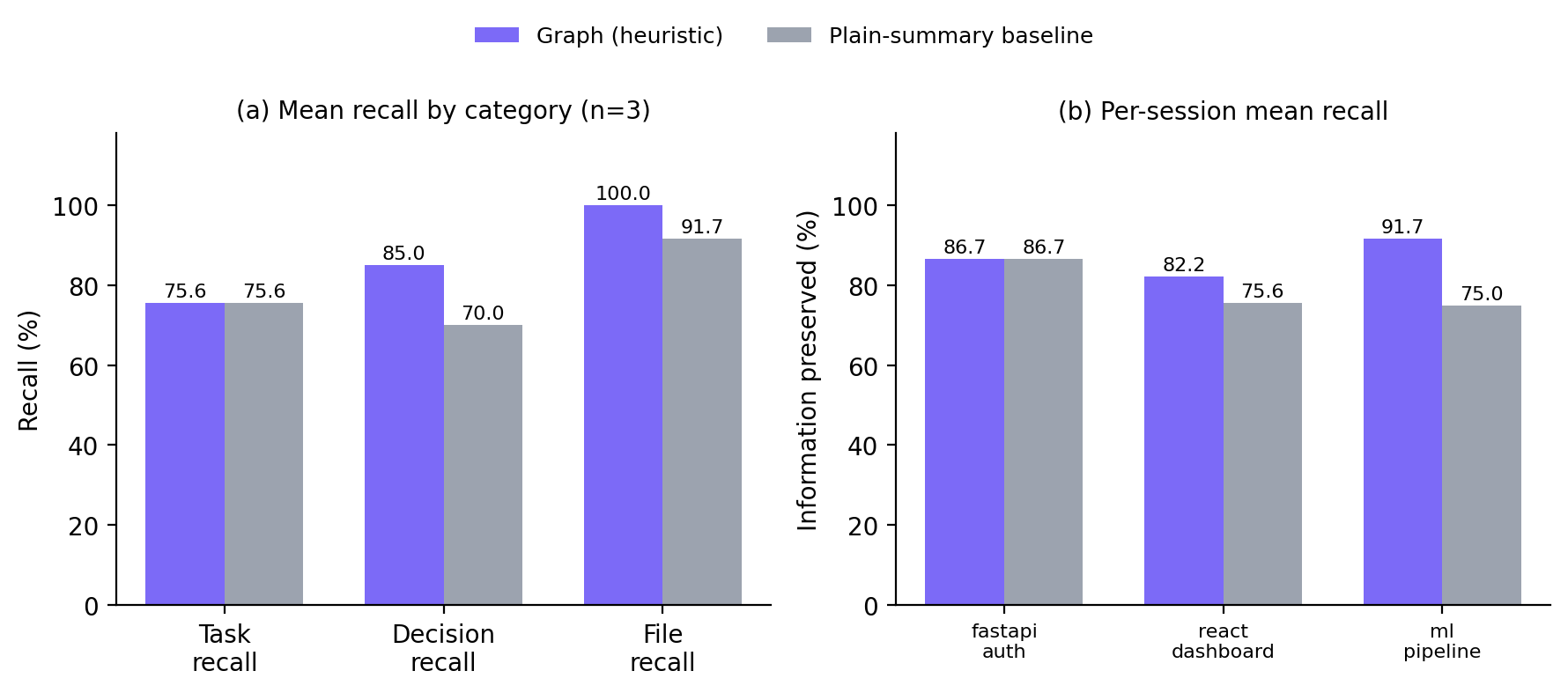}
  \caption{Graph extraction vs.\ plain-summary baseline
    ($n{=}3$, heuristic-only). (a)~Mean recall by category.
    (b)~Per-session mean recall
    (complement of Eq.~\eqref{eq:infoloss}). The advantage
    concentrates in decision recall.}
  \label{fig:recall}
\end{figure}

Table~\ref{tab:v3results} is the complete contents of the
results file; Figure~\ref{fig:recall} visualizes it.

\begin{table*}[t]
  \centering
  \caption{Complete per-session results (v0.3.1,
  heuristic-only; every measured value in
  \texttt{results\_v3\_code-0.3.1.json}).
  TR/DR/FR: task/decision/file recall.
  RT: standard-tier resume tokens (graph method).
  Nodes: extracted graph nodes.
  Time: full-session heuristic extraction (ms).}
  \label{tab:v3results}
  \renewcommand{\arraystretch}{1.25}
  \begin{tabular}{@{}lr rrr rrr rrr@{}}
    \toprule
    & & \multicolumn{3}{c}{\textbf{Graph (heuristic)}}
      & \multicolumn{3}{c}{\textbf{Plain-summary baseline}}
      & & & \\
    \cmidrule(lr){3-5}\cmidrule(lr){6-8}
    \textbf{Session} & \textbf{Turns} &
    \textbf{TR} & \textbf{DR} & \textbf{FR} &
    \textbf{TR} & \textbf{DR} & \textbf{FR} &
    \textbf{RT} & \textbf{Nodes} & \textbf{Time} \\
    \midrule
    \texttt{fastapi\_auth}    & 18 & 60.0\% & 100\% & 100\% & 60.0\% & 100\% & 100\%  & 302 & 27 & 529.9 \\
    \texttt{react\_dashboard} & 10 & 66.7\% &  80\% & 100\% & 66.7\% &  60\% & 100\%  & 260 & 27 &  23.6 \\
    \texttt{ml\_pipeline}     & 10 & 100\%  &  75\% & 100\% & 100\%  &  50\% &  75\%  & 201 & 17 &   8.1 \\
    \midrule
    \rowcolor{rowours}
    \textbf{Mean} & &
    \textbf{75.6\%} & \textbf{85.0\%} & \textbf{100\%} &
    \textbf{75.6\%} & \textbf{70.0\%} & \textbf{91.7\%} &
    \textbf{254} & \textbf{23.7} & \textbf{187.2} \\
    \bottomrule
  \end{tabular}
\end{table*}

Three observations, with the caution the sample demands.
\textbf{Task recall ties exactly} (75.6\% mean, identical per
session): under explicit completion markers, keyword scanning
recovers tasks as well as structured extraction.
\textbf{Decision recall separates the methods}: 85.0\% vs.\
70.0\% ($+$15.0~pp mean; $+$20 and $+$25~pp on the two
differing sessions)---despite the baseline's keyword list
overlapping the ground-truth vocabulary
(Section~\ref{sec:baseline}).
\textbf{File recall} is perfect for the graph method on all
three sessions vs.\ 91.7\% for the baseline; at these values a
ceiling effect is likely and harder sessions would be needed
to separate methods reliably.
By Eq.~\eqref{eq:infoloss}, mean information preserved is
86.9\% (graph) vs.\ 79.1\% (baseline); information loss 13.1\%
vs.\ 20.9\%.

\subsection{Resume Footprint and Latency}
\begin{figure}[t]
  \centering
  \includegraphics[width=\columnwidth]{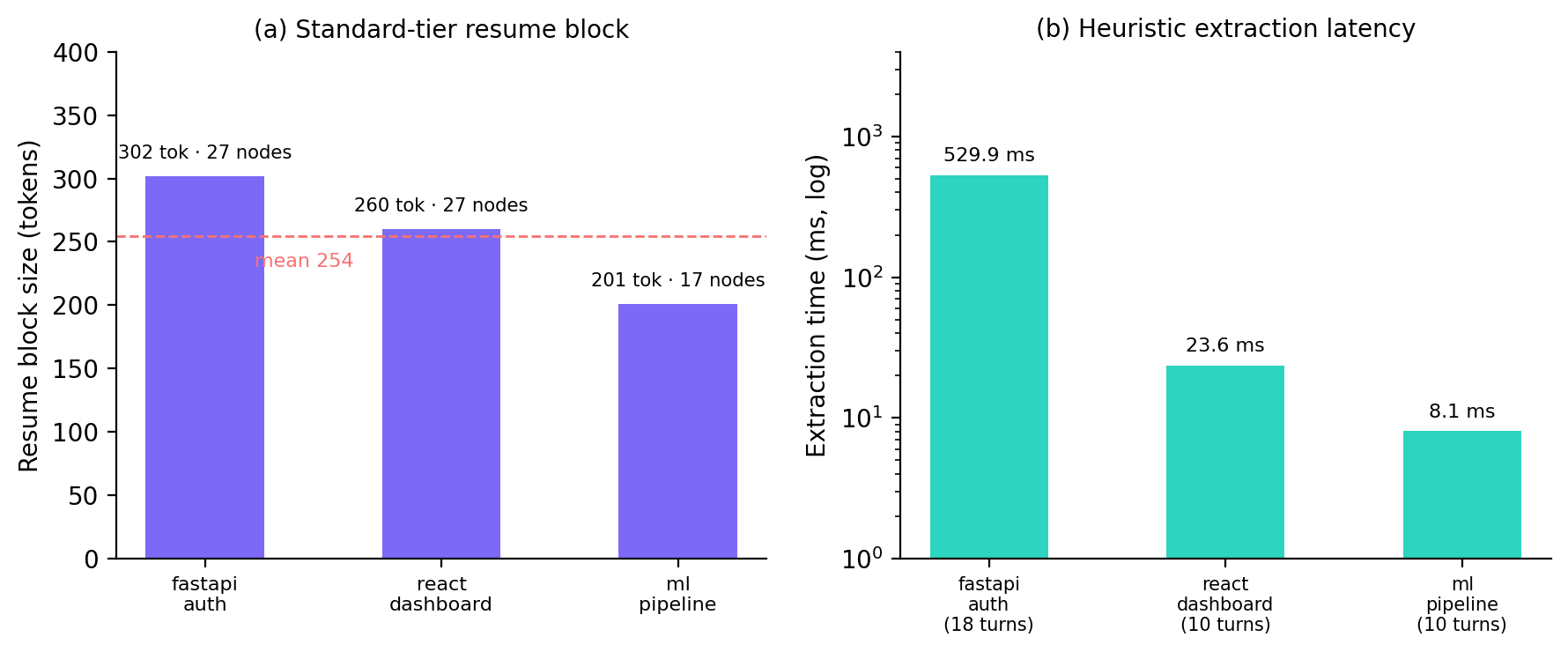}
  \caption{(a)~Standard-tier resume block size with extracted
    node counts. (b)~Full-session heuristic extraction latency
    (log scale); the 18-turn session is $22\times$ the 10-turn
    median.}
  \label{fig:footprint}
\end{figure}

Figure~\ref{fig:footprint} shows per-session resume sizes and
extraction times.
Standard-tier resumes measure 201--302 tokens (mean 254.3)
from graphs of 17--27 nodes; token efficiency
(Definition~\ref{def:eff}) is $\eta = 0.29$, $0.32$, $0.46$
(mean 0.35).
The resume replaces a multi-thousand-token transcript at
checkpoint time, but the results file records no comparable
footprint for the baseline, so \emph{no token-economy
comparison is claimed}.

Extraction latency spans two orders of magnitude: 8.1 and
23.6\,ms for the 10-turn sessions, 529.9\,ms for the 18-turn
session (mean 187.2\,ms, median 23.6\,ms).
All are below interactive thresholds---and extraction runs
post-response on the proxy path---but the $22\times$ jump on
the longest session is unexplained pending profiling
(Section~\ref{sec:limitations}).

\subsection{Context Window Recovery}
\begin{figure}[t]
  \centering
  \includegraphics[width=\columnwidth]{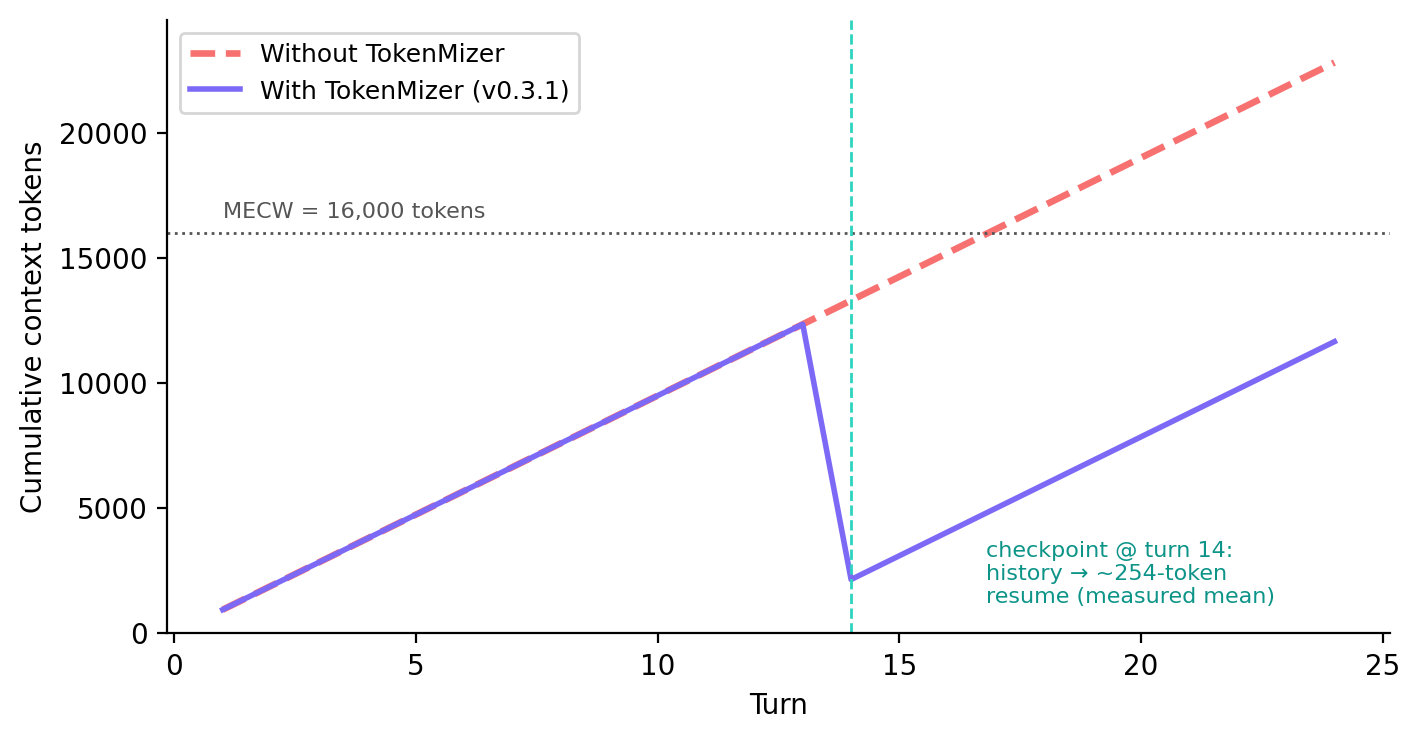}
  \caption{Cumulative context with and without TokenMizer for
    a hypothetical 950-token/turn session against a 16k MECW.
    The trajectory is analytic; the injected resume size
    (254 tokens) is the measured benchmark mean.}
  \label{fig:context}
\end{figure}

Figure~\ref{fig:context} illustrates the mechanism at the
measured mean resume size: at 950 tokens/turn against a 16k
MECW, the 85\% threshold triggers at turn~14, replacing
$\sim$13{,}300 accumulated tokens with a $\sim$254-token
resume, and the session continues past its overflow point.

\section{Discussion}
\label{sec:discussion}

\subsection{Decisions Are Where Structure Pays}
The $n{=}3$ pattern is small but coherent: flat text matches
structured extraction on tasks, nearly matches on files, and
loses on decisions.
This is the structural argument in miniature.
A keyword scan can report that ``redis'' occurs; it cannot
represent that Redis was \emph{chosen} for refresh-token
storage over the database---or, via decision transitions
(Section~\ref{sec:transitions}), that the choice was later
reversed and why.
A resume built from typed \texttt{DECISION} nodes with
supersession and invalidation preserves exactly the
information whose loss makes resumed sessions re-litigate
settled questions.

\subsection{What This Evaluation Cannot Show}
The results say nothing about implicit phrasing (all sessions
use explicit markers), strong baselines (an LLM summary would
likely beat the keyword scan), the compression and caching
subsystems (unmeasured in v0.3.1), or end-to-end benefit
(whether a 254-token resume actually improves a model's
continuation of the session).
Each is the natural next runner in the released benchmark
suite, which was structured so that additions do not change
the evaluation protocol.

\section{Limitations and Future Work}
\label{sec:limitations}

\textbf{L1: $n{=}3$ synthetic sessions.}
A smoke-level benchmark of released code, not an empirical
study; means over three points are orientation only.
The results file itself records the caveat
(``n=3 synthetic sessions; directional'').
A real-transcript suite is the highest-priority expansion.

\textbf{L2: Favorable phrasing; ceiling effects.}
All sessions use the explicit imperative style heuristics are
built for; file recall sits at 100\% and two decision-recall
values at or near 100\%, so the benchmark cannot discriminate
near the top.
v0.3.1 is unmeasured on implicit-phrasing sessions (academic
prose, consequence-style completions), where earlier code
versions performed substantially worse; do not extrapolate.

\textbf{L3: Single weak baseline.}
A keyword scan with a hand-specified list overlapping the
ground-truth vocabulary, reading the full transcript with no
budget.
LLM-summary, MemGPT~\cite{memgpt}, and retrieval baselines are
future work.

\textbf{L4: Author-constructed everything.}
The same person wrote the sessions, the ground truth, and the
extractor---a triple conflict that can inflate recall through
shared vocabulary.
Independent sessions and a second annotator are planned.

\textbf{L5: Precision unreported.}
The runner computes precision, but the v0.3.1 results file
persists recall only, so over-extraction is invisible here.
Persisting precision is a one-line planned schema change.

\textbf{L6: Compression, cache, hybrid path unmeasured.}
No v0.3.1 numbers exist for compression ratios, cache hit
rates, or LLM extraction; this paper accordingly claims none.

\textbf{L7: Latency scaling unprofiled.}
Whether the 18-turn session's 529.9\,ms reflects linear
message-count growth, superlinear graph effects, or fixed
warm-up cost is undetermined.

\textbf{L8: No end-to-end resumption metric.}
Entity recall is a proxy; the decisive measurement---does
injecting the resume improve downstream task
continuation?---has not been run.

\textbf{L9: No cross-session memory.}
Graphs are session-scoped; the SQLite backend would support
cross-session retrieval, but that layer is unimplemented.

\section{Conclusion}
\label{sec:conclusion}

TokenMizer v0.3.1 treats an LLM session as what it is: a
structured, revisable body of knowledge.
Its typed graph---14 node types, 7 edge types, an 8-state
lifecycle with supersession and invalidation, bitemporal
validity, and first-class decision-transition records---is
serialized by a tiered checkpoint system into token-budgeted
resume blocks, and is operational behind a streaming,
security-hardened transparent proxy with a dashboard,
visualization exports, and MCP tools that let agents manage
their own memory.

The evaluation is small by design and reproducible by
construction: three synthetic sessions, one results file, one
command.
Within that scope, graph memory ties a full-transcript keyword
baseline on task recall (75.6\%) and beats it on decision
recall (85.0\% vs.\ 70.0\%) and file recall (100\% vs.\
91.7\%), with 201--302-token resumes extracted in
8.1--529.9\,ms per session.
The supported claim is directional but pointed:
\emph{the entities that matter most for resumption---%
decisions---are the ones flat text loses and structure
keeps}.
Scaling that claim to real transcripts, strong baselines, and
end-to-end resumption metrics is the program the released
benchmark suite exists to carry.

\section*{Acknowledgment}
The author thanks the open-source communities behind FastAPI,
SQLite, sentence-transformers, and LLMLingua.

\balance

\appendix

\section{Fuzzy Matching Implementation}
\label{appendix:fuzzy}

The exact scoring functions from the benchmark runner
(\texttt{benchmarks/checkpoint\_accuracy/runner\_v2.py});
recall denominates over ground truth:

\begin{small}
\begin{verbatim}
def _fuzzy_match(a: str, b: str) -> bool:
    a, b = a.lower().strip(), b.lower().strip()
    if a in b or b in a:
        return True
    wa = set(re.findall(r'\w{3,}', a))
    wb = set(re.findall(r'\w{3,}', b))
    if not wa or not wb:
        return False
    shorter = wa if len(wa) <= len(wb) else wb
    return len(wa & wb) / len(shorter) >= 0.50

def _recall(extracted, expected) -> float:
    if not expected:
        return 1.0
    if not extracted:
        return 0.0
    matched = sum(
        1 for e in expected
        if any(_fuzzy_match(e, ex)
               for ex in extracted))
    return round(matched / len(expected), 3)
\end{verbatim}
\end{small}

\end{document}